\documentclass[conference]{IEEEtran}
\IEEEoverridecommandlockouts
\usepackage{cite}
\usepackage{amsmath,amssymb,amsfonts}
\usepackage{algorithmic}
\usepackage{graphicx}
\usepackage{textcomp}
\usepackage{xcolor}
\def\BibTeX{{\rm B\kern-.05em{\sc i\kern-.025em b}\kern-.08em
    T\kern-.1667em\lower.7ex\hbox{E}\kern-.125emX}}

\usepackage{bm}
\usepackage{mathtools}
\usepackage{url}
\usepackage{amsthm}

\newtheorem{defi}{Definition}
\newtheorem{prop}{Proposition}
\newtheorem{theo}{Theorem}

\newtheorem{rema}{Remark}

\newtheorem{assum}{Assumption}

\allowdisplaybreaks[4]

\begin{document}

\title{Variational Bayesian Methods for a Tree-Structured Stick-Breaking Process Mixture of Gaussians by Application of the Bayes Codes for Context Tree Models}

\author{\IEEEauthorblockN{Yuta Nakahara}
\IEEEauthorblockA{\textit{Center for Data Science} \\
\textit{Waseda University}\\
Tokyo, Japan \\
y.nakahara@waseda.jp}
}

\maketitle

\begin{abstract}
The tree-structured stick-breaking process (TS-SBP) mixture model is a non-parametric Bayesian model that can represent tree-like hierarchical structures among the mixture components. For TS-SBP mixture models, only a Markov chain Monte Carlo (MCMC) method has been proposed and any variational Bayesian (VB) methods has not been proposed. In general, MCMC methods are computationally more expensive than VB methods. Therefore, we require a large computational cost to learn the TS-SBP mixture model. In this paper, we propose a learning algorithm with less computational cost for the TS-SBP mixture of Gaussians by using the VB method under an assumption of finite tree width and depth. When constructing such VB method, the main challenge is efficient calculation of a sum over all possible trees. To solve this challenge, we utilizes a subroutine in the Bayes coding algorithm for context tree models. We confirm the computational efficiency of our VB method through an experiments on a benchmark dataset.
\end{abstract}


\section{Introduction}

Clustering is one of the major topics in machine learning and data science. In classical probabilistic models such as Gaussian mixture models, the number of mixture components had to be predetermined. Later, models like the Dirichlet process mixture model\cite{DP, DP2, CRP, SBP} were proposed. These models do not require predetermined numbers of mixture components. They sometimes called non-parametric Bayesian models. Further, in recent years, non-parametric Bayesian models with latent tree structures have been proposed\cite{BayesianHierarchicalClustering, TSSB, 2023SMC}. They do not only require predetermined numbers of mixture components but also represent hierarchical structures among the mixture components. The tree-structured stick-breaking process (TS-SBP) mixture model\cite{TSSB} is an example of such models, on which we focus in this paper.

For these Bayesian statistical models, there are two major algorithms used for learning their posterior distribution: Markov chain Monte Carlo (MCMC) methods and variational Bayesian (VB) methods (see, e.g., \cite{bishop}). The MCMC method can be applied to complicated models with various likelihood functions and prior distributions, and can compute the true posterior distribution after a sufficiently large number of iterations. However, the computational cost is generally higher than the VB method. The VB method cannot strictly represent the true posterior distribution because it restricts the class of probability distributions used to approximate the posterior distribution in advance. In particular, for non-parametric Bayesian models, the VB method requires setting a maximum number of possible mixture components in advance. However, it can obtain an approximate posterior distribution faster than the MCMC methods in general.

For the aformentioned non-parametric Bayesian models, various learning methods based on the MCMC methods and the VB methods have been proposed. For example, for the Dirichlet process mixture model, learning algorithms using both the MCMC methods\cite{DP_MCMC, DP_ParticleFilter, DP_SMC, SBP_GibbsSampling, SBP_GibbsSampling2, SBP_SliceSampling} and the VB methods\cite{DP_VB, DP_VB2} have been proposed. However, for the TS-SBP mixture model, only a learning algorithm using the MCMC method\cite{TSSB} has been proposed, which requires a large computational cost to learn. Therefore, in this study, we propose a learning algorithm with less computational cost for the TS-SBP mixture model by using the VB method.

In the VB methods, there are two main approaches to restricting the class of approximate posterior distributions (see, e.g., \cite{bishop}). The first assumes that the approximate posterior distribution can be factorized and each factor can be represented as a parametric distribution. The second assumes only that the approximate posterior distribution can be factorized, without assuming whether each distribution has a parametric representation. This study takes the second approach, which is based on weaker assumptions. In this approach, the update equations for the approximate posterior distribution may not be represented in any parametric form and may contain integrals and sums, depending on the likelihood function and prior distribution settings. In that case, the update equations may be computationally intractable. In particular, when constructing a VB method for the TS-SBP mixture model, it is necessary to efficiently calculate sums over all possible tree structures. This is the main challenge in this study.

To solve this challenge, this study utilizes a subroutine in the Bayes coding algorithm for context tree models\cite{CT_alg}. The Bayesian coding algorithm for context tree models is originally a model and algorithm for data compression in information theory. In this algorithm, by assuming a special prior distribution\cite{full_rooted_trees} for the tree structure that defines the context tree model, a subroutine can be used to calculate sums over all possible trees exactly and efficiently. We utilize this subroutine in deriving update equations for the VB method for the TS-SBP mixture model.

Finally, we confirm our VB method can learn hierarchical structures among mixture components and is computationally superior to the conventional MCMC method through experiments on a toy example and a benchmark dataset.

\section{Stochastic models}

\begin{figure}[htbp]
\centering
\includegraphics[width=\linewidth]{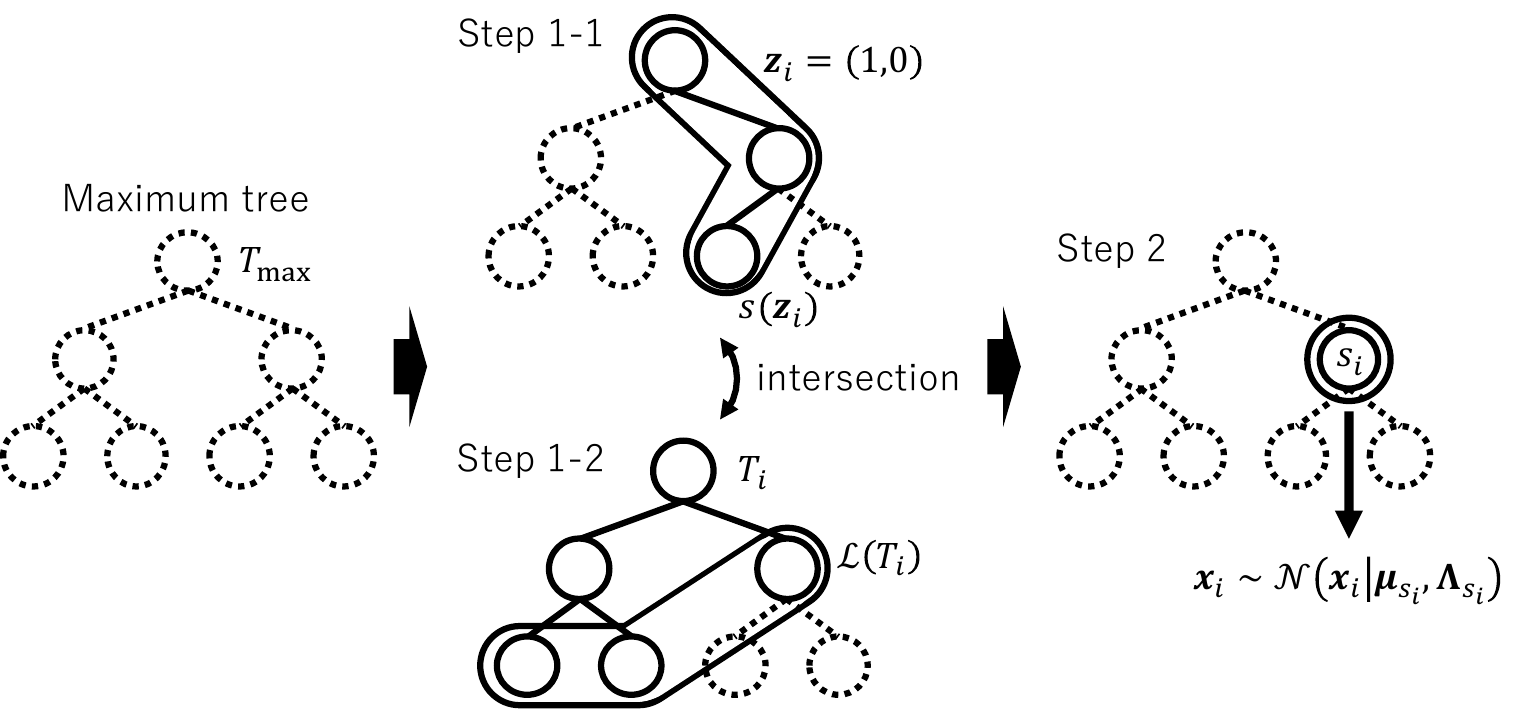}
\caption{An overview of the data generation process.}
\label{generative_model}
\end{figure}

In this section, we define the TS-SBP mixture of Gaussians in a manner different from \cite{TSSB}. This enables us to utilize a subroutine of the Bayes coding algorithm for the context tree models when deriving update equations in our VB method. In our definition, the $i$th data point $\bm x_i$ is independently generated according to the following steps (see also Fig.\ \ref{generative_model}).
\begin{itemize}
\item Step 1: latent node generation.
    \begin{itemize}
    \item Step 1-1: latent path generation.
    \item Step 1-2: latent subtree generation.
    \end{itemize}
\item Step 2: data point generation.
\end{itemize}

Let $K \in \mathbb{N}$ and $D \in \mathbb{N}$ be given constants. In Step 1, a node on a $K$-ary perfect\footnote{All inner nodes have exactly $K$ children and all leaf nodes have the same depth.} tree with depth $D$ is randomly selected. Let $T_\mathrm{max}$ denote the $K$-ary perfect tree. Step 1 consists of two independent substeps: Step 1-1 and Step 1-2. We describe each step in order.

In Step 1-1, a path from the root node to a leaf node of $T_\mathrm{max}$ is randomly selected in the following manner. First, we define some notations. Let $\mathcal{S}_\mathrm{max}$, $\mathcal{I}_\mathrm{max}$, and $\mathcal{L}_\mathrm{max}$ denote the set of all nodes, inner nodes, and leaf nodes of $T_\mathrm{max}$, respectively. Let $s_\lambda$ denote the root node of $T_\mathrm{max}$. The depth of any node $s \in \mathcal{S}_\mathrm{max}$ is denoted by $d_s$, e.g., $s_\lambda = 0$. We define $\mathrm{Ch}(s)$ as a set of child nodes of $s$ on $T_\mathrm{max}$.

We assume each inner node $s \in \mathcal{I}_\mathrm{max}$ has a routing parameter $\bm \pi_s \coloneqq (\pi_{s, s_\mathrm{ch}})_{s_\mathrm{ch} \in \mathrm{Ch}(s)} \in \mathbb{R}^K$ that satisfies $\sum_{s_\mathrm{ch} \in \mathrm{Ch}(s)} \pi_{s,s_\mathrm{ch}} = 1$, and we define its tuple $\bm \pi \coloneqq (\bm \pi_s)_{s \in  \mathcal{I}_\mathrm{max}}$. According to the routing parameter $\bm \pi$, a path from the root node $s_\lambda$ to a leaf node $s(\bm z_i)$ is randomly generated in the following manner. The path is determined by a latent variable $\bm z_i \coloneqq (z_{i,1}, z_{i,2}, \dots , z_{i,D})^\top \in \{ 0, 1, \dots , K-1 \}^D$. First, $z_{i,1}$ is generated according to $\bm \pi_{s_\lambda}$, i.e., $z_{i,1} = k$ holds with probability $\pi_{s_\lambda, s_k}$, where $s_k$ denotes the $k$th child node of $s_\lambda$. Then, $z_{i,2}$ is generated according to $\bm \pi_{s_k}$ in a similar manner. This procedure is repeated until we reach one of the leaf nodes of $T_\mathrm{max}$. Let $s(\bm z_i)$ denote the leaf node determined by this procedure. Thus, the probability distribution of $\bm z_i$ is represented as follows.
\begin{defi}
We define the probability distribution of $\bm z_i$ given $\bm \pi$ as follows:
\begin{align}
    p(\bm z_i | \bm \pi) \coloneqq \prod_{s \in \mathcal{I}_\mathrm{max}} \prod_{s_\mathrm{ch} \in \mathrm{Ch}(s)} \pi_{s,s_\mathrm{ch}}^{I\{ s_\mathrm{ch} \preceq s(\bm z_i) \}},
\end{align}
where $I \{ \cdot \}$ denotes the indicator function and $s_\mathrm{ch} \preceq s(\bm z_i)$ represents $s_\mathrm{ch}$ is an ancestor node of $s(\bm z_i)$ or equal to $s(\bm z_i)$.
\end{defi}

In Step 1-2, a latent subtree $T_i$ for the $i$th data point is generated according to the following procedure. Let $T_i$ denote a full (also called proper) subtree of $T_\mathrm{max}$, where $T_i$'s root node is $s_\lambda$. Each inner node of $T_i$ has exactly $K$ children.
Let $\mathcal{T}$ denote the set of all such subtrees of $T_\mathrm{max}$.
The set of all nodes, inner nodes, and leaf nodes of $T_i$ are denoted by $\mathcal{S}(T_i)$, $\mathcal{I}(T_i)$, and $\mathcal{L}(T_i)$, respectively.
Then, $T_i$ is generated according to the following probability distribution, which is used in text compression (e.g., \cite{CT_alg}) and mathematically summarized in \cite{full_rooted_trees}.

\begin{defi}[\cite{full_rooted_trees}]\label{TPrior}
\begin{align}
 p(T_i|\bm g) \coloneqq \prod_{s \in \mathcal{I}(T_i)} g_s \prod_{s' \in \mathcal{L}(T_i)} (1-g_{s'}), \label{DistributionTree}
\end{align}
where $g_s \in [0, 1]$ is a parameter representing an edge spreading probability of a node $s \in \mathcal{S}_\mathrm{max}$ and $\bm g$ denotes $(g_s)_{s \in \mathcal{S}_\mathrm{max}}$. For $s \in \mathcal{L}_\mathrm{max}$, we assume $g_s = 0$.
\end{defi}

\begin{rema}
Eq.\ \eqref{DistributionTree} satisfies the condition of the probability distribution over $\mathcal{T}$, i.e., $\sum_{T_i \in \mathcal{T}} p(T_i|\bm g) = 1$ holds. The meaning of \eqref{DistributionTree} is detailed in Fig.\ 2 in \cite{full_rooted_trees}.
Other properties have also been discussed in \cite{full_rooted_trees}.
\end{rema}

At the end of Step 1, we take the intersection of $\mathcal{L}(T_i)$ and the nodes on the path from $s_\lambda$ to $s(\bm z_i)$. Then, one of the nodes in $\mathcal{S}_\mathrm{max}$, i.e., a node $s$ that satisfies $(s \in \mathcal{L}(T_i)) \land (s \preceq s(\bm z_i))$, is uniquely determined. Let $s_i$ denote it.

In Step 2, a data point is generated and observed in the following manner. Let the dimension of the observed data be $p \in \mathbb{N}$. Let $\bm x_i = (x_{i,1}, x_{i,2}, \dots , x_{i,p})^\top \in \mathbb{R}^p$ denote the $i$th data point. We assume each node $s \in \mathcal{S}_\mathrm{max}$ has a mean vector $\bm \mu_s \in \mathbb{R}^p$ and a precision matrix $\bm \Lambda_s \in \mathbb{R}^{p \times p}$, which is assumed to be positive definite. We define the following tuples: $\bm \mu \coloneqq (\bm \mu_s)_{s \in \mathcal{S}_\mathrm{max}}$ and $\bm \Lambda \coloneqq (\bm \Lambda_s)_{s \in \mathcal{S}_\mathrm{max}}$. Given $T_i$, $\bm \mu$, $\bm \Lambda$, and $\bm z_i$, we assume the $i$th data point $\bm x_i$ is i.i.d.\ generated according to the following distribution.

\begin{defi}
We define the probability density function of $\bm x_i$ given $\bm z_i$, $T_i$, $\bm \mu$, and $\bm \Lambda$ as follows:
\begin{align}
     p(\bm x_i | \bm z_i, T_i, \bm \mu, \bm \Lambda) &\coloneqq \prod_{s \in \mathcal{L}(T_i)} \mathcal{N}(\bm x_i | \bm \mu_s, \bm \Lambda_s^{-1})^{I\{ s \preceq s(\bm z_i) \}}\\
     &= \mathcal{N}(\bm x_i | \bm \mu_{s_i}, \bm \Lambda_{s_i}^{-1}),
\end{align}
where $\mathcal{N}(\cdot)$ represents the probability density function of the Gaussian distribution.
\end{defi}

\begin{figure*}
    \centering
    \includegraphics[width=0.8\linewidth]{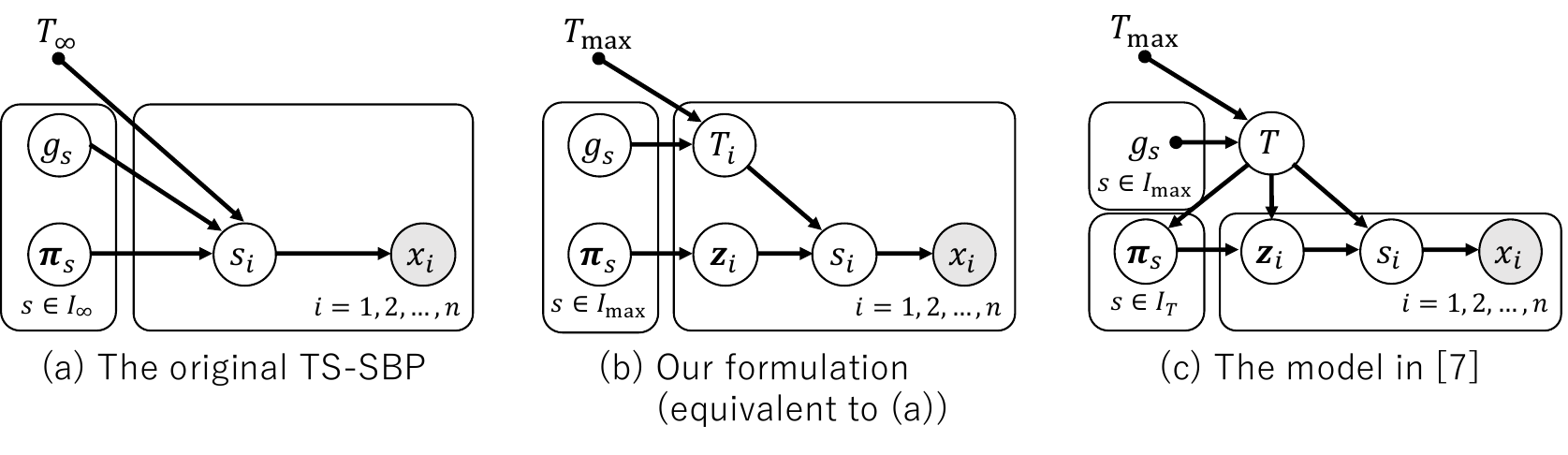}
    \caption{Comparison of our defined TS-SBP with the original TS-SBP and the model used in \cite{2023SMC} from the perspective of graphical models.}
    \label{model_comparison}
\end{figure*}

Note that this data generation process is equivalent to a truncated version of the TS-SBP in \cite{TSSB}. Let $s_i$ denote the node determined by $T_i$ and $\bm z_i$, i.e., $(s_i \in \mathcal{L}(T_i)) \land (s_i \preceq s(\bm z_i))$ holds, and $S_i$ denote the corresponding random variable on $\mathcal{S}_\mathrm{max}$.

First, using the graphical models in Fig.\ \ref{model_comparison}, we explain that our defined TS-SBP represents the original TS-SBP with finite tree width and depth in a different form of expression. Note that the parameters of $\bm x_i$ are omitted in Fig.\ \ref{model_comparison}. First, in the original TS-SBP, the probability of selecting node $s_i$ from an infinite-width and infinite-depth tree $T_\infty$ is directly determined by the product of branching probability $g_s$ and routing probability $\bm \pi_s$. (In \cite{TSSB}, $g_s$ corresponds to $1-\nu_{\bm \epsilon}$, and $\pi_{s_\mathrm{pa}, s}$ corresponds to $\varphi_{\bm \epsilon}$.) Therefore, the graphical model is as shown in Fig.\ \ref{model_comparison} (a). On the other hand, in our defined TS-SBP, the process of selecting node $s_i$ from a tree $T_\mathrm{max}$ with finite width and depth is described by separating it into the process of generating a subtree $T_i$ of $T_\mathrm{max}$ and the process of generating a path $\bm z_i$ on $T_\mathrm{max}$. As shown in Fig.\ \ref{generative_model}, once $T_i$ and $\bm z_i$ are determined, $s_i$ is uniquely determined. Therefore, the graphical model is as shown in Fig.\ \ref{model_comparison} (b). However, as we will prove it in the next theorem, if we marginalize over $T_i$ and $\bm z_i$, the probability of selecting node $s_i$ coincides with that of the original TS-SBP, except for the finite width and depth of the tree.

\begin{theo}\label{equivalence}
The probability distribution of $S_i$ over $\mathcal{S}_\mathrm{max}$ is represented as follows.
\begin{align}
    \mathrm{Pr} \{ S_i = s_i \} = (1-g_{s_i}) \pi_{(s_i)_\mathrm{pa},s_i} \prod_{s' \prec {s_i}} \pi_{s'_\mathrm{pa},s'} g_{s'},
\end{align}
where $s_\mathrm{pa}$ denotes the parent node of $s$ and we assume $\pi_{s_\mathrm{pa},s} = 1$ for $s=s_\lambda$. The above distribution is equivalent to a truncated version of the TS-SBP, where tree width and depth are $K$ and $D$, respectively.
\end{theo}

\proof:
\begin{align}
    &\mathrm{Pr} \{ S_i = s_i \} = \left( \sum_{T_i \in \mathcal{T}} I \{s_i \in \mathcal{L}(T_i) \} p(T_i|\bm g) \right) \nonumber \\
    &\qquad \times \left( \sum_{\bm z_i \in \{ 0, 1, \dots K-1 \}^D} I \{ s_i \preceq s(\bm z_i) \} p(\bm z_i | \bm \pi) \right) \\
    &= \left( (1-g_{s_i}) \prod_{s' \prec {s_i}} g_{s'} \right) \left( \pi_{(s_i)_\mathrm{pa},s_i} \prod_{s' \prec {s_i}} \pi_{s'_\mathrm{pa},s'} \right) \\
    &= (1-g_{s_i}) \pi_{(s_i)_\mathrm{pa},s_i} \prod_{s' \prec {s_i}} \pi_{s'_\mathrm{pa},s'} g_{s'}. \label{tssb_equivalent}
\end{align}
In the second equation, we used Theorem 2 of \cite{full_rooted_trees}. Eq.\ \eqref{tssb_equivalent} is equivalent to Eq.\ (2) in \cite{TSSB} except that the tree width and depth are limited. It is because $\mathrm{Pr} \{ S_i = s_i \}$, $g_{s_i}$, $1-g_{s_i}$, and $\pi_{(s_i)_\mathrm{pa},s_i}$ correspond to $\pi_{\bm \epsilon}$, $1-\nu_{\bm \epsilon}$, $\nu_{\bm \epsilon}$, and $\varphi_{\bm \epsilon}$ in \cite{TSSB}, respectively. \hfill $\Box$

Next, we clarify the difference between our TS-SBP and the tree-structured mixture distribution used in \cite{2023SMC} from the perspective of graphical models because that model is very similar to ours, and VB methods have also been derived for it. The most significant difference lies in whether the subtree $T$ of $T_\mathrm{max}$ is placed on the plate for $i=1, 2, \dots , n$ or not. In the model used in \cite{2023SMC}, $T$ is not placed on the plate, so it is generated only once throughout the data generation process. Then, node $s_i$ is selected not from $T_\mathrm{max}$, but from the leaf nodes $\mathcal{L}_{T}$ of the pre-generated subtree $T$.

\section{Variational Bayesian methods}

We assume $\{ \bm x_i \}_{i=1}^n$ is generated and observed according to the model described in the previous section. Let $\bm x$ denote $\{ \bm x_i \}_{i=1}^n$. We assume a prior distribution for $\bm \pi$, $\bm g$, $\bm \mu$, and $\bm \Lambda$ and estimate the posterior distribution for $\{ \bm z_i \}_{i=1}^n$, $\{ T_i \}_{i=1}^n$, $\bm \pi$, $\bm g$, $\bm \mu$, $\bm \Lambda$ from $\bm x$. Hereafter, let $\bm z$ denote $\{ \bm z_i \}_{i=1}^n$ and $\bm T$ denote $\{ T_i \}_{i=1}^n$. Unfortunately, it is difficult to exactly calculate the posterior distribution $p(\bm z, \bm T, \bm \pi, \bm g, \bm \mu, \bm \Lambda | \bm x)$. To approximate it, only MCMC methods have been proposed in \cite{TSSB}, but any VB mehtods have not been proposed yet. Herein, we propose a VB method.
To reduce the computational complexity, we impose the following assumptions on the prior distribution.

\subsection{Prior distributions}

\begin{assum}\label{PiPrior}
Given $\bm \alpha_s \in \mathbb{R}_{>0}^K$ for all $s \in \mathcal{I}_\mathrm{max}$, we assume the following probability distribution for $\bm \pi$, which is also known as Dirichlet tree distributions \cite{dirichlet_tree}.
\begin{align}
     p(\bm \pi ) \coloneqq \prod_{s \in \mathcal{I}_\mathrm{max}} \mathrm{Dir}(\bm \pi_s | \bm \alpha_s),
\end{align}
where $\mathrm{Dir}(\cdot)$ denotes a probability density function of the Dirichlet distribution.
\end{assum}

\begin{assum}
Given $a_s \in \mathbb{R}_{>0}$ and $b_s \in \mathbb{R}_{>0}$ for all $s \in \mathcal{I}_\mathrm{max}$, we assume the following probability distribution for $\bm g$, which is known as a conjugate prior of $p(T_i|\bm g)$ \cite{full_rooted_trees}.
\begin{align}
    p(\bm g) \coloneqq \prod_{s \in \mathcal{I}_\mathrm{max}} \mathrm{Beta}(g_s | a_s, b_s),
\end{align}
where $\mathrm{Beta}(\cdot)$ denotes a probability density function of the beta distribution. Note that $g_s$ for $s \in \mathcal{L}_\mathrm{max}$ is not a random variable but a constant equal to $0$.
\end{assum}

\begin{assum}\label{LambdaPrior}
Given a real number $\nu_s > p-1$ and a positive definite matrix $\bm W_s \in \mathbb{R}^{p \times p}$ for all $s \in \mathcal{S}_\mathrm{max}$, we assume the following probability distribution for $\bm \Lambda$.
\begin{align}
     p(\bm \Lambda ) &\coloneqq \prod_{s \in \mathcal{S}_\mathrm{max}} \mathcal{W}(\bm \Lambda_s | \nu_s, \bm W_s),
\end{align}
where $\mathcal{W}(\cdot)$ denotes a probability density function of the Wishart distribution.
\end{assum}

\begin{assum}\label{MuPrior}
Given a real number $u > p-1$ and a positive definite matrix $\bm V \in \mathbb{R}^{p \times p}$, we assume the following probability distribution for $\bm \mu$ with an additional parameter $\bm L \in \mathbb{R}^{p\times p}$.
\begin{align}
     p(\bm L ) &\coloneqq \mathcal{W}(\bm L | u, \bm V), \\
     p(\bm \mu | \bm L ) &\coloneqq \prod_{s \in \mathcal{S}_\mathrm{max}} \mathcal{N}(\bm \mu_s | \bm \mu_{s_\mathrm{pa}}, \bm L^{-1}),
\end{align}
where $s_\mathrm{pa}$ denotes the parent node of $s$. For the root node $s_\lambda$, we assume $\bm \mu_{(s_\lambda)_\mathrm{pa}} \coloneqq \bm m_{s_\lambda}$ is given as a hyperparameter.
\end{assum}

By Assumption \ref{MuPrior}, mixture components whose means are close to each other tend to be descendant nodes of a common node.

\subsection{Overview of variational Bayesian methods}

In the VB method, the posterior distribution $p(\bm z, \bm T, \bm \pi,$ $\bm g, \bm \mu, \bm \Lambda, \bm L | \bm x)$ is approximated by a distribution $q(\bm z, \bm T, \bm \pi,$ $\bm g, \bm \mu, \bm \Lambda, \bm L)$ called \emph{variational distribution} (see, e.g., \cite{bishop}). We assume our variational distribution fulfills the following factorization property.

\begin{assum}\label{factorization}
We assume the following factorization.
\begin{align}
    &q(\bm z, \bm T, \bm \pi, \bm g, \bm \mu, \bm \Lambda, \bm L) = q(\bm g) q(\bm L) \prod_{i=1}^n q(\bm z_i) \prod_{i=1}^n q(T_i) \nonumber \\
    &\qquad \times \prod_{s \in \mathcal{I}_\mathrm{max}} q(\bm \pi_s) \prod_{s \in \mathcal{S}_\mathrm{max}} q(\bm \mu_s) \prod_{s \in \mathcal{S}_\mathrm{max}} q(\bm \Lambda_s). \label{strong_factorization}
\end{align}
\end{assum}

Note that we do not assume any parametric form on each factor of the above variational distribution. In this situation, it is known that the optimal variational distribution that minimizes the Kullback-Leibler divergence $\mathrm{KL}(q(\bm z, \bm T, \bm \pi, \bm g, \bm \mu, \bm \Lambda, \bm L) || p(\bm z, \bm T, \bm \pi, \bm g,$ $\bm \mu, \bm \Lambda, \bm L | \bm x))$, which is equivalent to maximize the \emph{variational lower bound}, fulfills following equations (see, e.g., \cite{bishop}).
\begin{align}
&\ln q^* (\bm z_i) \!=\! \mathbb{E}_{\backslash \bm z_i} \left[ \ln p(\bm x, \bm z, \bm T, \bm \pi, \bm g, \bm \mu, \bm \Lambda, \bm L) \right] \!+\! \mathrm{const.}, \label{q_star_z}\\
&\ln q^* (T_i) \!=\! \mathbb{E}_{\backslash T_i} \left[ \ln p(\bm x, \bm z, \bm T, \bm \pi, \bm g, \bm \mu, \bm \Lambda, \bm L) \right] \!+\! \mathrm{const.}, \label{q_star_T}\\
&\ln q^* (\bm \pi_s) \!=\! \mathbb{E}_{\backslash \bm \pi_s} \left[ \ln p(\bm x, \bm z, \bm T, \bm \pi, \bm g, \bm \mu, \bm \Lambda, \bm L) \right] \!+\! \mathrm{const.}, \label{q_star_pi} \\
&\ln q^* (\bm g) \!=\! \mathbb{E}_{\backslash \bm g} \left[ \ln p(\bm x, \bm z, \bm T, \bm \pi, \bm g, \bm \mu, \bm \Lambda, \bm L) \right] \!+\! \mathrm{const.}, \label{q_star_g} \\
&\ln q^* (\bm \mu_s) \!=\! \mathbb{E}_{\backslash \bm \mu_s} \! \left[ \ln p(\bm x, \bm z, \bm T, \bm \pi, \bm g, \bm \mu, \bm \Lambda, \bm L) \right] \!+\! \mathrm{const.}, \label{q_star_mu}\\
&\ln q^* (\bm \Lambda_s) \!=\! \mathbb{E}_{\backslash \bm \Lambda_s} \! \left[ \ln p(\bm x, \bm z, \bm T, \bm \pi, \bm g, \bm \mu, \bm \Lambda, \bm L) \right] \!+\! \mathrm{const.},\! \label{q_star_lambda}\\
&\ln q^* (\bm L) \!=\! \mathbb{E}_{\backslash \bm L} \left[ \ln p(\bm x, \bm z, \bm T, \bm \pi, \bm g, \bm \mu, \bm \Lambda, \bm L) \right] \!+\! \mathrm{const.} \label{q_star_L}
\end{align}
where $\mathbb{E}_{\backslash (\star)}$ means the expectation for all the latent variables except $(\star)$ with respect to the variational distribution $q$.

However, $q^* (\bm z_i)$, $q^* (T_i)$, $q^* (\bm \pi_s)$, $q^* (\bm g)$, $q^* (\bm \mu_s)$, $q^*(\bm \Lambda_s)$ and $q^*(\bm L)$ depend on each other. Therefore, we update them in turn from an initial value until the convergence by using Eqs.\ \eqref{q_star_z} to \eqref{q_star_L} as updating formulas. However, these expectations include integrals and sums, which may be infeasible in general. Particularly, these expectations include the sums with respect to all the subtrees of $T_\mathrm{max}$, which require doubly exponential cost for the depth of $T_\mathrm{max}$. To calculate these sums, we have to derive an efficient parametric representation of Eq.\ \eqref{q_star_T} only from the factorization assumption in Assumption \ref{factorization}. To mathematically derive it, we utilize a subroutine used in the Bayes coding algorithm for the context tree models.

\subsection{Update of $q(\bm \pi_s)$, $q(\bm \mu_s)$, $q(\bm \Lambda_s)$, $q(\bm L)$, $q(\bm g)$, and $q(\bm z_i)$}

For $\bm \pi_s$, $\bm \mu_s$, $\bm \Lambda_s$, $\bm L$, $\bm g$, and $\bm z_i$, we assumed locally conjugate prior distributions. Therefore, $q(\bm \pi_s)$, $q(\bm \mu_s)$, $q(\bm \Lambda_s)$, $q(\bm L)$, $q(\bm g)$, and $q(\bm z_i)$ have the same form as the prior distributions.

\begin{prop}\label{prop:q_other_T}
There exist parameters $\hat{\bm \alpha}_s \in \mathbb{R}_{>0}^K$, $\hat{\bm m}_s \in \mathbb{R}^p$, $\hat{\bm L}_s \in \mathbb{R}^{p \times p}$, $\hat{\nu}_s \in \mathbb{R}$, $\hat{\bm W}_s \in \mathbb{R}^{p \times p}$, $\hat{u} \in \mathbb{R}$, $\hat{\bm V} \in \mathbb{R}^{p \times p}$, $\hat{a}_s \in \mathbb{R}_{>0}$, $\hat{b}_s \in \mathbb{R}_{>0}$, and $\hat{\pi}_{i,s,s_\mathrm{ch}} \in \mathbb{R}_{>0}$ such that $q(\bm \pi_s)$, $q(\bm \mu_s)$, $q(\bm \Lambda_s)$, $q(\bm L)$, $q(\bm g)$, and $q(\bm z_i)$ have the following representation.
\begin{align}
    q(\bm \pi_s) &= \mathrm{Dir}(\bm \pi_s | \hat{\bm \alpha}_s),\\
    q(\bm \mu_s) &= \mathcal{N}(\bm \mu_s | \hat{\bm m}_s, \hat{\bm L}_s^{-1}), \label{q_mu}\\
    q(\bm \Lambda_s) &= \mathcal{W}(\bm \Lambda_s | \hat{\nu}_s, \hat{\bm W}_s), \label{q_Lambda}\\
    q(\bm L) &= \mathcal{W}(\bm L | \hat{u}, \hat{\bm V}), \label{q_L}\\
    q(\bm g) &= \prod_{s \in \mathcal{I}_\mathrm{max}} \mathrm{Beta}(g_s | \hat{a}_s, \hat{b}_s), \label{q_g}\\
    q(\bm z_i) &= \prod_{s \in \mathcal{I}_\mathrm{max}} \prod_{s_\mathrm{ch} \in \mathrm{Ch}(s)} \hat{\pi}_{i,s,s_\mathrm{ch}}^{I\{ s_\mathrm{ch} \preceq s(\bm z_i) \}}.
\end{align}
\end{prop}

This proposition is almost straightforwardly proved by using basic theorems in Bayesian statistics and theorems in \cite{full_rooted_trees}. The proof and specific updating formulas of the parameters are detailed in Appendix \ref{appendix_q_other_T}.

Based on Proposition \ref{prop:q_other_T}, we define some quantities in advance. First, for any $s$, let $q(s \preceq s(\bm z_i))$ denote the probability that the event $\{ \bm z_i \mid s \preceq s(\bm z_i) \}$ occurs under $q(\bm z_i)$, i.e.,
\begin{align}
    q(s \preceq s(\bm z_i)) \coloneqq & \mathbb{E}_{q(\bm z_i)}[I\{ s \preceq s(\bm z_i) \}] \label{q_s_s_z} \\
    =& \prod_{s' \preceq s} \hat{\pi}_{i,s'_\mathrm{pa},s'}.
\end{align}
In addition, we define the following quantities.
\begin{align}
    \ln \phi_{i,s} &\coloneqq q(s \preceq s(\bm z_i)) \mathbb{E}_{q(\bm \mu_s, \bm \Lambda_s)} [\ln \mathcal{N}(\bm x_i | \bm \mu_s, \bm \Lambda_s^{-1})], \label{phi_def} \\
    \ln \tilde{g}_s &\coloneqq \mathbb{E}_{q(g_s)}[\ln g_s], \label{g_tilde_def} \\
    \ln \tilde{g}_s^c &\coloneqq \mathbb{E}_{q(g_s)}[\ln (1-g_s)]. \label{g_tilde_c_def}
\end{align}

\subsection{Update of $q(T_i)$}

As described before, integrals and sums in Eq. \eqref{q_star_T} may not be feasible in general. To solve this, we utilize a subroutine in the Bayes coding algorithm for the context tree models. This subroutine enables us to calculate the expectation in Eq.\ \eqref{q_star_T} without any approximation, and the following parametric representation is derived only from the factorization property assumed in Assumption \ref{factorization}.

\begin{theo}\label{theo:q_T}
The updating formula for $q(T_i)$ is given as follows:
\begin{align}
     q(T_i) = \prod_{s \in \mathcal{I}(T_i)} \hat{g}_{i,s} \prod_{s' \in \mathcal{L}(T_i)} (1-\hat{g}_{i,s'}), \label{q_T}
\end{align}
where $\hat{g}_{i,s} \in [0,1]$ is obtained as follows:
\begin{align}
    \hat{g}_{i,s} \coloneqq
    \begin{cases}
        \frac{\tilde{g}_s \prod_{s_\mathrm{ch} \in \mathrm{Ch}(s)} \rho_{i,s_\mathrm{ch}}}{\rho_{i,s}}, & s \in \mathcal{I}_\mathrm{max},\\
        0, & s \in \mathcal{L}_\mathrm{max}.
    \end{cases}\label{g_hat}
\end{align}
Here, $\rho_{i,s}$ is recursively defined in the following manner.
\footnote{To calculate $\rho_{i,s}$, we should calculate $\ln \rho_{i,s}$ by using the logsumexp function rather than directly calculating $\rho_{i,s}$.}
\begin{align}
    \rho_{i,s} \! \coloneqq \!
    \begin{cases}
        \tilde{g}_s^c \phi_{i,s} + \tilde{g}_s \prod_{s_\mathrm{ch} \in \mathrm{Ch}(s)} \rho_{i,s_\mathrm{ch}},\! & s \in \mathcal{I}_\mathrm{max},\\
        \phi_{i,s}, & s \in \mathcal{L}_\mathrm{max}.
    \end{cases} \label{ln_rho}
\end{align}
\end{theo}

The proof of this theorem is in Appendix \ref{appendix_q_T}. It should be noted that Eqs. \eqref{g_hat} and \eqref{ln_rho} correspond to the Bayes coding algorithm for context tree sources, e.g., Eqs. (12) and (9) in \cite{CT_alg}, respectively.

\subsection{Initialization}

In this paper, we use the following initialization of the variational distribution. We initialize the parameters of $q(\bm T, \bm \pi, \bm g, \bm \mu, \bm \Lambda, \bm L)$ and start updating from $q(\bm z)$. First, we deterministically initialize the parameters other than $\hat{\bm m}_s$ for any node $s$ and its child node $s_\mathrm{ch}$ as follows: $\hat{u} = u$, $\hat{\bm V} = \bm V$
    $\hat{a}_s = a_s$,
    $\hat{b}_s = b_s$,
    $\hat{g}_s = \frac{a_s}{a_s + b_s}$,
    $\hat{\alpha}_{s,s_\mathrm{ch}} = \alpha_{s,s_\mathrm{ch}}$,
    $\hat{\bm L}_s = u \bm V$,
    $\hat{\nu}_s = \nu_s$, and 
    $\hat{\bm W}_s = \bm W_s$.
Next, $\hat{\bm m}_s$ is randomly and recursively assigned from the root node as follows:
\begin{align}
    \hat{\bm m}_{s_\lambda} &= \frac{1}{n} \sum_{i=1}^n \bm x_i, \\
    \hat{\bm m}_s &\sim \mathcal{N}(\hat{\bm m}_s | \hat{\bm m}_{s_\mathrm{pa}}, (u\bm V)^{-1} ).
\end{align}
Therefore, the mean $\hat{\bm m}_s$ of each node $s$ is centered on the parent's mean $\hat{\bm m}_{s_\mathrm{pa}}$.

\section{Experiments}\label{sec:experiments}

\subsection{Toy example}

\begin{figure}[tbp]
\centering
\includegraphics[width=\linewidth]{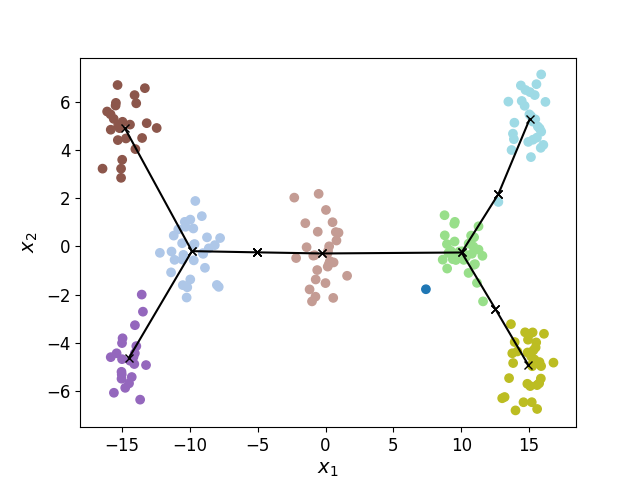}
\caption{The input data and the estimated tree structure of the means of the mixture components.}
\label{result_synthetic}
\end{figure}

\begin{figure*}[tbp]
\centering
\includegraphics[width=0.8\linewidth]{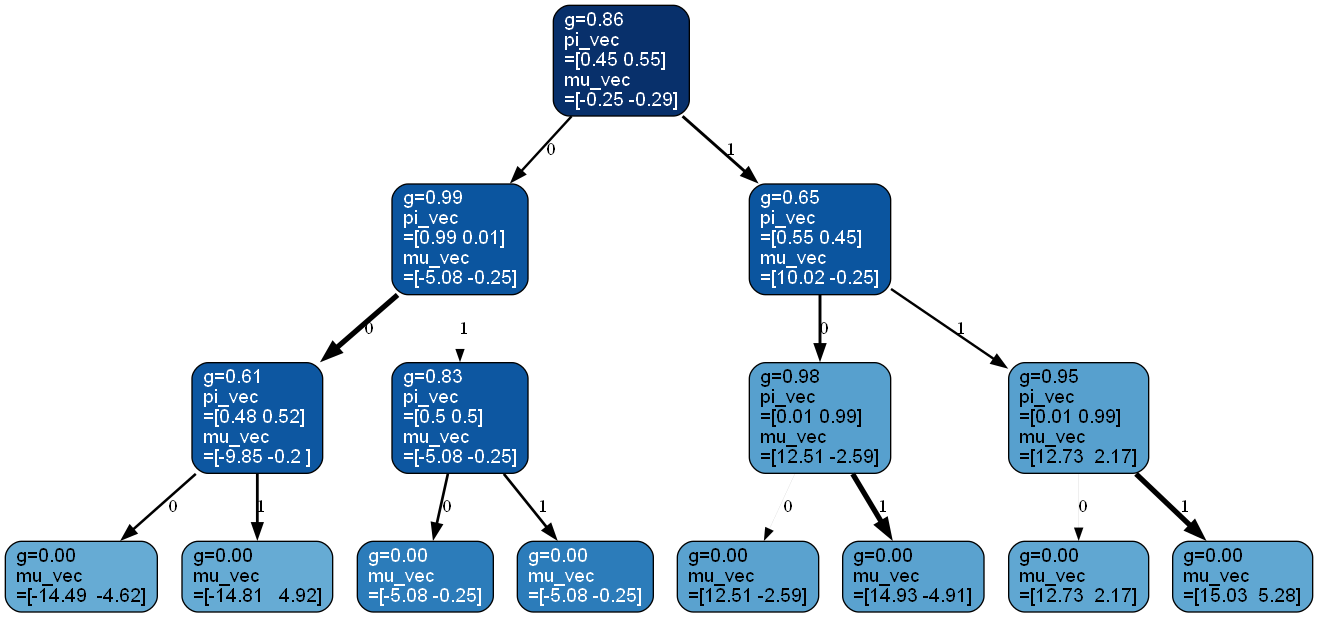}
\caption{The TS-SBP mixture of Gaussians estimated from the data shown in Fig.\ \ref{result_synthetic}.}
\label{estimated_TSSBPGMM}
\end{figure*}

In this section, we show an experimental result on synthetic data. The data are generated from a Gaussian mixture model. The means of mixture components are $[-15,-5]^\top$, $[-15,5]^\top$, $[-10,0]^\top$, $[0,0]^\top$, $[10,0]^\top$, $[15,-5]^\top$, and $[15,5]^\top$, and the covariance matrices are all the identity matrix $\bm I$. The mixing probability is uniform. Figure \ref{result_synthetic} shows the scatter plots of the generated data. The sample size is 200.

The constants of the TS-SBP mixture of Gaussians are assumed to be $p=2$, $K=2$, and $D=3$. Therefore, we have at most $15$ mixture components. We set the hyperparameters as follows: $a_s=3$, $b_s=1$, $\alpha_{s,s_\mathrm{ch}}=1/2$, $\bm m_{s_\lambda}=[0,0]^\top$, $u=5$, $\bm V = \bm I / 10$, $\nu_s=2$, and $\bm W_s = \bm I / 5$ for any $s$ and $s_\mathrm{ch}$. The maximum number of iterations is assumed to be 400. Initial values of variational distributions are randomly generated 100 times by the procedure in the previous section. The variational distribution that shows the largest variational lower bound is used for parameter estimation.

Figures \ref{result_synthetic} and \ref{estimated_TSSBPGMM} show the estimated model. In Fig.\ \ref{result_synthetic}, the cross `$\times$' represents $\hat{\bm m}_s$ for each node $s$. The color circle shows the MAP node for each data point. All the parameters are estimated by the expectations of the variational distributions. As shown in Fig.\ \ref{result_synthetic}, mixture components close to each other tend to be children of a common inner node as expected. 

\subsection{CIFAR-100 image dataset}

\begin{figure*}[tbp]
\centering
\includegraphics[width=0.9\linewidth]{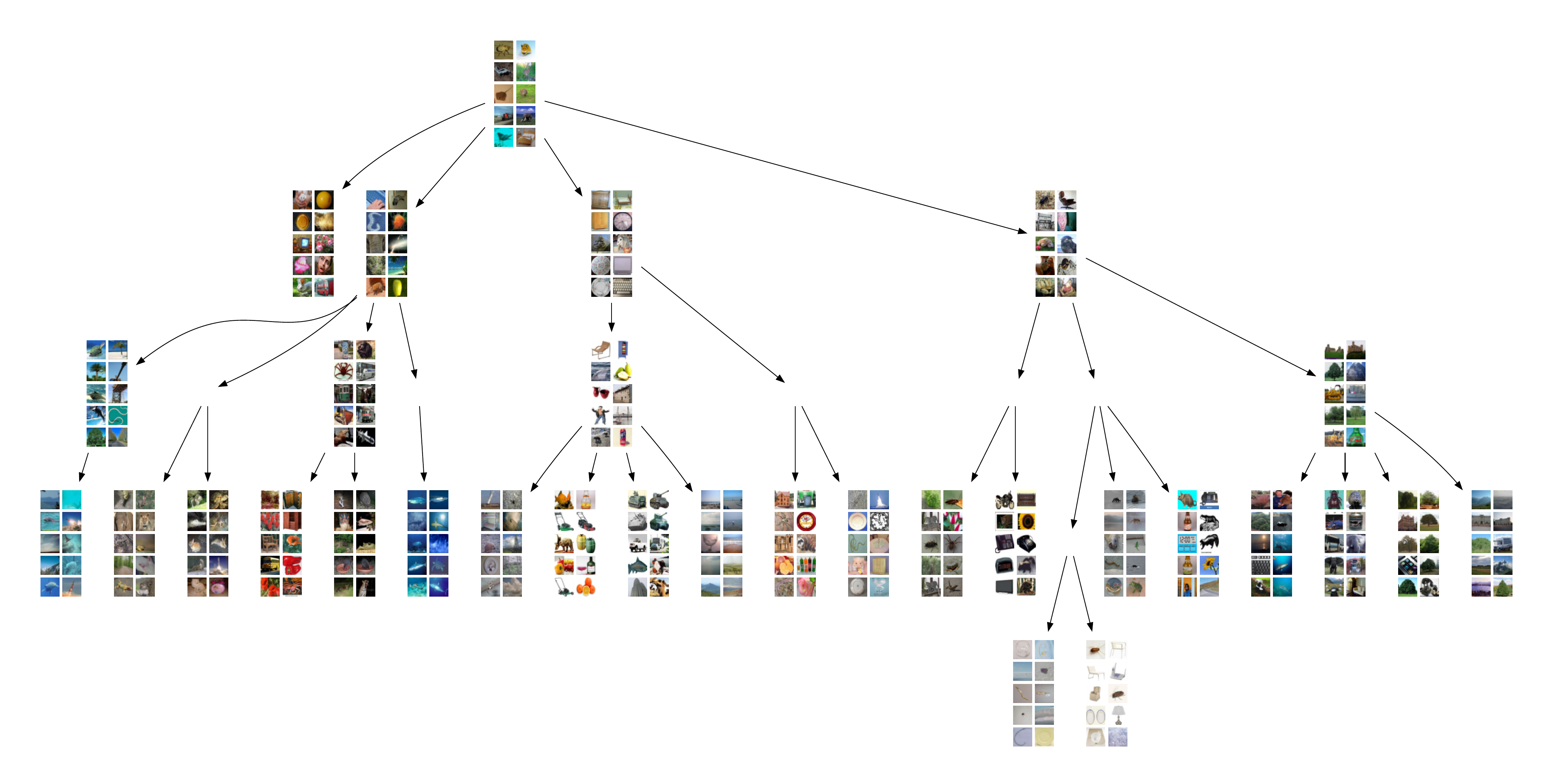}
\caption{The images are placed at nodes with the maximum a posteriori probability. We only show the node with at least 50 images, and we show the 10 images with highest probabilities at each node.}
\label{cifar100_result}
\end{figure*}

The purpose of this experiment is to apply our VB method to the same data (CIFAR-100) used in \cite{TSSB}, in order to compare our VB method with the MCMC method in \cite{TSSB}. However, since the source code for the algorithm proposed in \cite{TSSB} is not publicly available and the preprocessing is not so detailed, a direct comparison is difficult. Therefore, we will only make a rough comparison of computational complexity under the same data scale. The input data consist of 50,000 images from CIFAR-100. In \cite{TSSB}, the data dimension $p$ was reduced to 256 dimensions through preprocessing using deep learning\cite{preprocessingDNN}, but in this study, we reduced it to 256 dimensions using principal component analysis. Thus, the size of the input data is the same for both our study and \cite{TSSB}. The hyperparameters in our method were set as follows. The number of child nodes $K$ for each node was set to 4, and the maximum depth $D$ of the tree $T_\mathrm{max}$ was set to 4. The hyperparameters for the prior distribution were set as follows: $a_s=100\times (0.1)^{d_s}$, $b_s=1$, $\bm \alpha_s =[1, 0.1, 0.01, 0.001]^\top$, $\bm m_{s_\lambda}=\bm 0$, $u=512$, $\bm V = 10 \bm S / u$, $\nu_s=256$, and $\bm W_s = 1000 \bm I / \nu_s$ for any $s$, where $d_s$ represents the depth of node $s$, and $\bm S$ represents the sample covariance matrix. The number of iterations for the VB method was set to 100, and the initial values were reset 5 times. The result with the largest variational lower bound at the 100th iteration was adopted.

The results are shown in Figure \ref{cifar100_result}. In the MCMC method of \cite{TSSB}, it took about 3 minutes to sample from all latent variables using Python on a single-core workstation, and this was repeated 4000 times. Therefore, it took about 200 hours in total. With our VB method, it took about 90 seconds per iteration, and we obtained the results in about 13 hours in total, including resetting the initial values. Limiting the width and depth of the tree and using the VB method reduced the computation time. The details of the machine used are as follows. OS: Windows 11, CPU: Intel(R) Xeon(R) w7-3445, Memory: 64GB, Language: Python. Explicit parallel processing was not implemented, and no GPU was used. Looking at the clustering results, it appears that similar images are gathered in each node, and the nodes close to each other have a similar characteristics. This is the expected result from the definition of the model. The clustering in \cite{TSSB} appears to be more fine-grained, but this may be due to differences in preprocessing.

\section{Conclusion}

We proposed a learning algorithm for the TS-SBP mixture of Gaussians by using the VB method. That required less computational cost than the conventional MCMC method under an assumption of finite tree width and depth. The main challenge in deriving the updating equations in the VB method for the TS-SBP mixture models was efficient calculation of a sum over all possible trees. To solve this challenge, we utilized a subroutine in the Bayes coding algorithm for context tree models. Finally, we confirmed the computational efficiency of our VB method through an experiment on a benchmark dataset.


\appendices

\section{Proof of Proposition \ref{prop:q_other_T}}\label{appendix_q_other_T}

In tho following, we divide Proposition \ref{prop:q_other_T} into six propositions for each of $q(\bm \pi_s)$, $q(\bm \mu_s)$, $q(\bm \Lambda_s)$, $q(\bm L)$, $q(\bm g)$, and $q(\bm z_i)$ and prove them in order.

\subsubsection{Update of $q(\bm \pi_s)$}

First, we define the following notation.
\begin{align}
     N_s \coloneqq \sum_{i=1}^n q(s \preceq s(\bm z_i)), \label{n_s}
\end{align}
where $q(s \preceq s(\bm z_i))$ is defined in \eqref{q_s_s_z}. Then, the following proposition holds.

\begin{prop}\label{prop:q_pi}
The updating formula for $q(\bm \pi_s)$ is given as $q(\bm \pi_s) = \mathrm{Dir}(\bm \pi_s | \hat{\bm \alpha}_s)$, where each element of $\hat{\bm \alpha}_s \in \mathbb{R}_{>0}^K$ is obtained as follows:
\begin{align}
    \hat{\alpha}_{s,s_\mathrm{ch}} \coloneqq \alpha_{s,s_\mathrm{ch}} + N_{s_\mathrm{ch}}. \label{alpha_hat}
\end{align}
\end{prop}

\proof Calculating \eqref{q_star_pi}, we obtain the following equation.
\begin{align}
    \ln q(\bm \pi_s) &= \sum_{s_\mathrm{ch} \in \mathrm{Ch}(s)} \left( \sum_{i=1}^n q(s_\mathrm{ch} \preceq s(\bm z_i)) \right) \ln \pi_{s,s_\mathrm{ch}} \nonumber \\
    &\quad + \ln \mathrm{Dir}(\bm \pi_s | \bm \alpha_s) + \mathrm{const.}, \label{q_pi}
\end{align}
Therefore, Proposition \ref{prop:q_pi} holds. \hfill $\Box$

\subsection{Update of $q(\bm \mu_s)$}

First, for any $s$, let $q(s \in \mathcal{L}(T_i))$ denote the probability that the event $\{ T_i \mid s \in \mathcal{L}(T_i) \}$ occurs under $q(T_i)$,\footnote{We can calculate it as $q(s \in \mathcal{L}(T_i)) = (1-\hat{g}_{i,s}) \prod_{s' \prec s} \hat{g}_{i,s'}$ by using Theorem 2 in \cite{full_rooted_trees}.} i.e.,
\begin{align}
    q(s \in \mathcal{L}(T_i)) \coloneqq \mathbb{E}_{q(T_i)}[I\{ s \in \mathcal{L}(T_i) \}]. \label{leaf_prob}
\end{align}
Next, we define the following notations.
\begin{align}
    \tilde{N}_s &\coloneqq \sum_{i=1}^n q(s \in \mathcal{L}(T_i)) q(s \preceq s(\bm z_i)), \label{n_tilde_s}\\
    \bar{\bm x}_s &\coloneqq \frac{1}{\tilde{N}_s}  \sum_{i=1}^n q(s \in \mathcal{L}(T_i)) q(s \preceq s(\bm z_i)) \bm x_i. \label{x_bar_s}
\end{align}
Then, the following proposition holds. 

\begin{prop}\label{prop:q_mu}
The updating formula for $q(\bm \mu_s)$ is given as $q(\bm \mu_s) = \mathcal{N}(\bm \mu_s | \hat{\bm m}_s, \hat{\bm L}_s^{-1})$, where $\hat{\bm m}_s \in \mathbb{R}^p$ and $\hat{\bm L}_s \in \mathbb{R}^{p \times p}$ are obtained as follows:
\begin{align}
    \hat{\bm L}_s \coloneqq &\begin{cases}
        \tilde{N}_s \mathbb{E}_{q(\bm \Lambda_s)}[\bm \Lambda_s] + (K+1) \mathbb{E}_{q(\bm L)}[\bm L], & s \in \mathcal{I}_\mathrm{max}, \\
        \tilde{N}_s \mathbb{E}_{q(\bm \Lambda_s)}[\bm \Lambda_s] + \mathbb{E}_{q(\bm L)}[\bm L], & s \in \mathcal{L}_\mathrm{max},
    \end{cases}\label{L_hat}\\
    \hat{\bm m}_s \coloneqq &\begin{cases}
        \hat{\bm L}_s^{-1} \bigl\{ \tilde{N}_s \mathbb{E}_{q(\bm \Lambda_s)}[\bm \Lambda_s] \bar{\bm x}_s + \mathbb{E}_{q(\bm L)}[\bm L] \\
        \qquad \quad \times (\hat{\bm m}_{s_\mathrm{pa}} + \sum_{s_\mathrm{ch \in \mathrm{Ch}(s)}} \hat{\bm m}_{s_\mathrm{ch}}) \bigr\}, & s \in \mathcal{I}_\mathrm{max}, \\
        \hat{\bm L}_s^{-1} \bigl\{ \tilde{N}_s \mathbb{E}_{q(\bm \Lambda_s)}[\bm \Lambda_s] \bar{\bm x}_s \\
        \qquad \quad + \mathbb{E}_{q(\bm L)}[\bm L] \hat{\bm m}_{s_\mathrm{pa}} \bigr\}, & s \in \mathcal{L}_\mathrm{max}.
    \end{cases}\label{m_hat}
\end{align}
\end{prop}

\proof We prove Proposition \ref{prop:q_mu} only for $s \in \mathcal{I}_\mathrm{max}$ because those for $s \in \mathcal{L}_\mathrm{max}$ will be proved similarly. Calculating \eqref{q_star_mu}, we obtain the following equation.
\begin{align}
    &\ln q(\bm \mu_s) \nonumber \\
    &= \sum_{i=1}^n \mathbb{E}_{q(T_i)} \Biggl[ \sum_{s \in \mathcal{L}(T_i)} q(s \preceq s(\bm z_i)) \nonumber \\
    &\qquad \qquad \qquad \qquad \quad \times \mathbb{E}_{q(\bm \Lambda_s)}[ \ln \mathcal{N}(\bm x_i | \bm \mu_s, \bm \Lambda_s^{-1}) ] \Biggr] \nonumber \\
    &\quad + \mathbb{E}_{q(\bm \mu_{s_\mathrm{pa}}, \bm L)} [\ln \mathcal{N}(\bm \mu_s | \bm \mu_{s_\mathrm{pa}}, \bm L^{-1})] \nonumber \\
    &\quad + \sum_{s_\mathrm{ch} \in \mathrm{Ch}(s)} \mathbb{E}_{q(\bm \mu_{s_\mathrm{ch}}, \bm L)} [\ln \mathcal{N}(\bm \mu_{s_\mathrm{ch}} | \bm \mu_s, \bm L^{-1})] + \mathrm{const.}
\end{align}

Here, we can calculate the expectation for $q(T_i)$ as follows:
\begin{align}
    & \mathbb{E}_{q(T_i)} \left[ \sum_{s \in \mathcal{L}(T_i)} q(s \preceq s(\bm z_i)) \mathbb{E}_{q(\bm \Lambda_s)}[ \ln \mathcal{N}(\bm x_i | \bm \mu_s, \bm \Lambda_s^{-1}) ] \right] \nonumber \\
    &= \mathbb{E}_{q(T_i)} \Biggl[ \sum_{s \in \mathcal{S}_\mathrm{max}} I\{ s \in \mathcal{L}(T_i) \} q(s \preceq s(\bm z_i)) \nonumber \\
    &\qquad \qquad \qquad \qquad \quad \times \mathbb{E}_{q(\bm \Lambda_s)}[ \ln \mathcal{N}(\bm x_i | \bm \mu_s, \bm \Lambda_s^{-1}) ] \Biggr] \label{E_T0}\\    
    &= \sum_{s \in \mathcal{S}_\mathrm{max}} \mathbb{E}_{q(T_i)} [ I\{ s \in \mathcal{L}(T_i) \} ] q(s \preceq s(\bm z_i)) \nonumber \\
    &\qquad \qquad \qquad \qquad \quad \times \mathbb{E}_{q(\bm \Lambda_s)}[ \ln \mathcal{N}(\bm x_i | \bm \mu_s, \bm \Lambda_s^{-1}) ]. \label{E_T}
\end{align}
This technique is often used in the proof of Proposition \ref{prop:q_other_T}. Note that $q(s \in \mathcal{L}(T_i))$ was defined by $\mathbb{E}_{q(T_i)} [ I\{ s \in \mathcal{L}(T_i) \} ]$.

After this, we can prove Proposition \ref{prop:q_mu} in a similar manner to the proof of the conjugate property of Gaussian distributions for Gaussian likelihoods with given covariance matrices. \hfill $\Box$

\subsection{Update of $q(\bm \Lambda_s)$}

First, we define the following notation.
\begin{align}
    &\bm S_s \coloneqq \frac{1}{\tilde{N}_s}  \sum_{i=1}^{n} q(s \in \mathcal{L}(T_i)) q(s \preceq s(\bm z_i)) \nonumber \\
    &\qquad \qquad \qquad \quad \times (\bm x_i - \hat{\bm m}_s)(\bm x_i - \hat{\bm m}_s)^\top. \label{S_s}
\end{align}
Then, the following proposition holds.

\begin{prop}\label{prop:q_Lambda}
The updating formula for $q(\bm \Lambda_s)$ is given as $q(\bm \Lambda_s) = \mathcal{W}(\bm \Lambda_s | \hat{\nu}_s, \hat{\bm W}_s)$, where $\hat{\nu}_s \in \mathbb{R}$ and $\hat{\bm W}_s \in \mathbb{R}^{p \times p}$ are obtained as follows:
\begin{align}
    \hat{\nu}_s &\coloneqq \nu_s + \tilde{N}_s, \label{nu_hat}\\
    \hat{\bm W}_s^{-1} &\coloneqq \bm W_s^{-1} + \tilde{N}_s (\bm S_s + \hat{\bm L}_s^{-1}). \label{w_hat}
\end{align}
\end{prop}

\proof In a similar manner to \eqref{E_T0} and \eqref{E_T}, we transform \eqref{q_star_lambda} as follows.
\begin{align}
    \ln q(\bm \Lambda_s) =& \sum_{i=1}^n \mathbb{E}_{q(T_i)} [ I\{ s \in \mathcal{L}(T_i) \} ] q(s \preceq s(\bm z_i)) \nonumber \\
    &\qquad \times \mathbb{E}_{q(\bm \mu_s)}[ \ln \mathcal{N}(\bm x_i | \bm \mu_s, \bm \Lambda_s^{-1}) ] \nonumber \\
    &+ \ln \mathcal{W}(\bm \Lambda_s | \nu_s, \bm W_s) + \mathrm{const.}
\end{align}
Then, in a similar manner to the proof of the conjugate property of Wishart distributions for Gaussian likelihoods with given means, Proposition \ref{prop:q_Lambda} holds. \hfill $\Box$

\subsection{Update of $q(\bm L)$}

\begin{prop}\label{prop:q_L}
The updating formula for $q(\bm L)$ is given as $q(\bm L) = \mathcal{W}(\bm L | \hat{u}, \hat{\bm V})$, where $\hat{u} \in \mathbb{R}$ and $\hat{\bm V} \in \mathbb{R}^{p \times p}$ are obtained as follows:
\begin{align}
    \hat{u} &\coloneqq u + |\mathcal{S}_\mathrm{max}|, \label{u_hat}\\
    \hat{\bm V}^{-1} &\coloneqq \bm V^{-1} + \hat{\bm L}_{s_\lambda}^{-1} + (\hat{\bm m}_{s_\lambda} - \bm m_{s_\lambda})(\hat{\bm m}_{s_\lambda} - \bm m_{s_\lambda})^\top, \nonumber \\
    &\qquad + \sum_{s \in \mathcal{S}_\mathrm{max} \backslash \{ s_\lambda \}} \Bigl( \hat{\bm L}_s^{-1} + \hat{\bm L}_{s_\mathrm{pa}}^{-1} \nonumber \\
    &\qquad \qquad \quad + (\hat{\bm m}_{s_\lambda} - \hat{\bm m}_{s_\mathrm{pa}})(\hat{\bm m}_{s_\lambda} - \hat{\bm m}_{s_\mathrm{pa}})^\top \Bigr), \label{v_hat}
\end{align}
where $| \cdot |$ represents the number of elements in a set, i.e., $|\mathcal{S}_\mathrm{max}|$ means the total number of the nodes in $T_\mathrm{max}$. 
\end{prop}

\proof Calculating \eqref{q_star_L}, we obtain the following equation.
\begin{align}
    \ln q(\bm L) &= \mathbb{E}_{q(\bm \mu_{s_\lambda})}[ \ln \mathcal{N}(\bm \mu_{s_\lambda} | \bm m_{s_\lambda}, \bm L^{-1})] \nonumber \\
    &\quad + \sum_{s \in \mathcal{S}_\mathrm{max} \setminus \{ s_\lambda \}} \mathbb{E}_{q(\bm \mu_s, \bm \mu_{s_\mathrm{pa}})}[ \ln \mathcal{N}(\bm \mu_s | \bm \mu_{s_\mathrm{pa}}, \bm L^{-1})] \nonumber \\
    &\quad + \ln \mathcal{W}(\bm L | u, \bm V) + \mathrm{const}.
\end{align}
Then, in a similar manner to the proof of the conjugate property of Wishart distributions for Gaussian likelihoods with given means, Proposition \ref{prop:q_L} holds. \hfill $\Box$

\subsection{Update of $q(\bm g)$}\label{update_of_q_g}

First, for any $s$, let $q(s \in \mathcal{I}(T_i))$ denote the probability that the event $\{ T_i \mid s \in \mathcal{I}(T_i) \}$ occurs under $q(T_i)$ in a similar manner to $q(s \in \mathcal{L}(T_i))$,\footnote{We can calculate it as $q(s \in \mathcal{I}(T_i)) = \hat{g}_{i,s} \prod_{s' \prec s} \hat{g}_{i,s'}$ by using Theorem 2 in \cite{full_rooted_trees}.} i.e.,
\begin{align}
    q(s \in \mathcal{I}(T_i)) \coloneqq \mathbb{E}_{q(T_i)}[I\{ s \in \mathcal{I}(T_i) \}]. \label{inner_prob}
\end{align}
Then, the following proposition holds.

\begin{prop}\label{prop:q_g}
The updating formula for $q(\bm g)$ is given as $q(\bm g) = \prod_{s \in \mathcal{I}_\mathrm{max}} \mathrm{Beta}(g_s | \hat{a}_s, \hat{b}_s)$, where $\hat{a}_s \in \mathbb{R}_{>0}$ and $\hat{b}_s \in \mathbb{R}_{>0}$ are obtained as follows:
\begin{align}
    \hat{a}_s &\coloneqq a_s + \sum_{i=1}^n q(s \in \mathcal{I}(T_i)), \label{a_hat}\\
    \hat{b}_s &\coloneqq b_s + \sum_{i=1}^n q(s \in \mathcal{L}(T_i)). \label{b_hat}
\end{align}
\end{prop}

\proof Calculating \eqref{q_star_g}, we obtain the following equation.
\begin{align}
    &\ln q(\bm g) \nonumber \\
    &= \sum_{i=1}^n \mathbb{E}_{q(T_i)} \left[ \sum_{s \in \mathcal{I}(T_i)} \ln g_s + \sum_{s \in \mathcal{L}(T_i) \setminus \mathcal{L}_\mathrm{max}} \ln (1-g_s) \right] \nonumber \\
    &\qquad +\sum_{s \in \mathcal{I}_\mathrm{max}} \ln \mathrm{Beta}(g_s | a_s, b_s) + \mathrm{const.}\\
    &= \sum_{s \in \mathcal{I}_\mathrm{max}} \Biggl\{ \left( \sum_{i=1}^n \mathbb{E}_{q(T_i)}[I\{ s \in \mathcal{I}(T_i)] \right) \ln g_s \nonumber \\
    &\qquad \qquad \quad + \left( \sum_{i=1}^n \mathbb{E}_{q(T_i)}[I\{ s \in \mathcal{L}(T_i)] \right) \ln ( 1 - g_s ) \nonumber \\
    &\qquad \qquad \quad + \ln \mathrm{Beta}(g_s | a_s, b_s) \Biggr\} + \mathrm{const.}
\end{align}
Here, we used a technique similar to that used in \eqref{E_T0} and \eqref{E_T}. Note that $q(s \in \mathcal{I}(T_i))$ and $q(s \in \mathcal{L}(T_i))$ were defined by $\mathbb{E}_{q(T_i)} [ I\{ s \in \mathcal{I}(T_i) \} ]$ and $\mathbb{E}_{q(T_i)} [ I\{ s \in \mathcal{L}(T_i) \} ]$, respectively. Then, in a similar manner to the conjugate property of beta distributions for likelihoods of Bernoulli distributions, Proposition \ref{prop:q_g} holds. \hfill $\Box$

\subsection{Update of $q(\bm z_i)$}

\begin{prop}\label{prop:q_z}
The updating formula for $q(\bm z_i)$ is as follows:
\begin{align}
     q(\bm z_i) = \prod_{s \in \mathcal{I}_\mathrm{max}} \prod_{s_\mathrm{ch} \in \mathrm{Ch}(s)} \hat{\pi}_{i,s,s_\mathrm{ch}}^{I\{ s_\mathrm{ch} \preceq s(\bm z_i) \}}, \label{q_z}
\end{align}
where $\hat{\pi}_{i,s,s_\mathrm{ch}} \in \mathbb{R}_{>0}$ is defined as
\begin{align}
     \hat{\pi}_{i,s,s_\mathrm{ch}} \coloneqq \frac{\xi_{i,s,s_\mathrm{ch}}}{\sum_{s_\mathrm{ch} \in \mathrm{Ch}(s)} \xi_{i,s,s_\mathrm{ch}}},
\end{align}
and $\xi_{i,s,s_\mathrm{ch}} \in \mathbb{R}_{>0}$ is recursively defined as follows:\footnote{To calculate $\ln \sum_{s' \in \mathrm{Ch}(s_\mathrm{ch})} \xi_{i,s_\mathrm{ch},s'}$, we should use the logsumexp function.}
\begin{align}
    \ln \xi_{i,s,s_\mathrm{ch}} \coloneqq \begin{cases}
        \mathbb{E}_{q(\bm \pi_s)}[\ln \pi_{s, s_\mathrm{ch}}] + \ln \zeta_{i,s_\mathrm{ch}} \\
        \quad + \ln \sum_{s' \in \mathrm{Ch}(s_\mathrm{ch})} \xi_{i,s_\mathrm{ch},s'} , & s_\mathrm{ch} \in \mathcal{I}_\mathrm{max}, \\
        \mathbb{E}_{q(\bm \pi_s)}[\ln \pi_{s, s_\mathrm{ch}}] + \ln \zeta_{i,s_\mathrm{ch}}, & s_\mathrm{ch} \in \mathcal{L}_\mathrm{max}.
    \end{cases}
\end{align}
Here, $\zeta_{i,s}$ is defined as follows:
\begin{align}
    \ln \zeta_{i,s} \coloneqq q(s \in \mathcal{L}(T_i)) \mathbb{E}_{q(\bm \mu_s, \bm \Lambda_s)}[\ln \mathcal{N}(\bm x_i | \bm \mu_s, \bm \Lambda_s^{-1})] \label{ln_zeta}
\end{align}
\end{prop}

\proof In a similar manner to \eqref{E_T0} and \eqref{E_T}, we transform \eqref{q_star_z} as follows.
\begin{align}
    &\ln q (\bm z_i) \nonumber \\
    &= \sum_{s \in \mathcal{S}_\mathrm{max}} \mathbb{E}_{q(T_i)} [ I \{ s \in \mathcal{L}(T_i) \} ] I\{ s \preceq s(\bm z_i) \} \nonumber \\
    &\qquad \qquad \quad  \times \mathbb{E}_{q(\bm \mu_s, \bm \Lambda_s)}[ \ln \mathcal{N}(\bm x_i | \bm \mu_s, \bm \Lambda_s^{-1})] \nonumber \\
    &\quad \!+\! \sum_{s \in \mathcal{I}_\mathrm{max}} \sum_{s_\mathrm{ch} \in \mathrm{Ch}(s)} \! I\{ s_\mathrm{ch} \preceq s(\bm z_i) \} \mathbb{E}_{q(\bm \pi_s)}[\ln \pi_{s, s_\mathrm{ch}}] + \mathrm{const.} \\
    &= \ln \zeta_{i,s_\lambda} + \sum_{s \in \mathcal{I}_\mathrm{max}} \sum_{s_\mathrm{ch} \in \mathrm{Ch}(s)} I\{ s_\mathrm{ch} \preceq s(\bm z_i) \} \nonumber \\
    &\qquad \qquad \qquad \quad \times ( \mathbb{E}_{q(\bm \pi_s)}[\ln \pi_{s, s_\mathrm{ch}}] + \ln \zeta_{i,s_\mathrm{ch}}) + \mathrm{const.}, \label{root_is_const}
\end{align}
where $\ln \zeta_{i,s}$ is defined in \eqref{ln_zeta}. Note that the first term of \eqref{root_is_const} is independent of $\bm z_i$.

Next, we substitute the definitions of $\hat{\pi}_{i,s,s_\mathrm{ch}}$ and $\xi_{i,s,s_\mathrm{ch}}$ into the logarithm of \eqref{q_z} and bring it back to \eqref{root_is_const}. First, we show another representation of the logarithm of the right-hand side of \eqref{q_z} with the notation $s_\mathrm{pa}$ that represents the parent node of $s$.
\begin{align}
    &\sum_{s \in \mathcal{I}_\mathrm{max}} \sum_{s_\mathrm{ch} \in \mathrm{Ch}(s)} I \{ s_\mathrm{ch} \preceq s(\bm z_i) \} \ln \hat{\pi}_{i,s,s_\mathrm{ch}} \nonumber \\
    &= \sum_{s \in \mathcal{I}_\mathrm{max} \setminus \{ s_\lambda \}} I \{ s \preceq s(\bm z_i) \} \ln \hat{\pi}_{i,s_\mathrm{pa},s} \nonumber \\
    &\qquad + \sum_{s \in \mathcal{L}_\mathrm{max}} I \{ s \preceq s(\bm z_i) \} \ln \hat{\pi}_{i,s_\mathrm{pa},s}
\end{align}
Next, we substitute the definitions of $\hat{\pi}_{i,s,s_\mathrm{ch}}$ and $\xi_{i,s,s_\mathrm{ch}}$.
\begin{align}
    &\sum_{s \in \mathcal{I}_\mathrm{max} \setminus \{ s_\lambda \}} I \{ s \preceq s(\bm z_i) \} \Biggl( \mathbb{E}_{q(\bm \pi_{s_\mathrm{pa}})}[\ln \pi_{s_\mathrm{pa}, s}] + \ln \zeta_{i,s} \nonumber \\
    &\qquad \qquad + \ln \sum_{s_\mathrm{ch} \in \mathrm{Ch}(s)} \xi_{i,s,s_\mathrm{ch}} - \ln \sum_{s' \in \mathrm{Ch}(s_\mathrm{pa})} \xi_{i,s_\mathrm{pa},s'} \Biggr) \nonumber \\
    &\quad + \sum_{s \in \mathcal{L}_\mathrm{max}} I \{ s \preceq s(\bm z_i) \} \Biggl( \mathbb{E}_{q(\bm \pi_{s_\mathrm{pa}})}[\ln \pi_{s_\mathrm{pa}, s}] \nonumber \\
    &\qquad \qquad \qquad \qquad + \ln \zeta_{i,s} - \ln \sum_{s \in \mathrm{Ch}(s_\mathrm{pa})} \xi_{i,s_\mathrm{pa},s} \Biggr).
\end{align}
Here, $\ln \sum_{s_\mathrm{ch} \in \mathrm{Ch}(s)} \xi_{i,s,s_\mathrm{ch}}$ for most $s$ is canceled like a telescoping sum, and only $\ln \sum_{s' \in \mathrm{Ch}(s_\lambda)} \xi_{i,s_\lambda,s'}$ will remained. Further, we represent the sum in the original form. Then, we obtain the following formula.
\begin{align}
    & \sum_{s \in \mathcal{I}_\mathrm{max}} \sum_{s_\mathrm{ch} \in \mathrm{Ch}(s)} I \{ s_\mathrm{ch} \preceq s(\bm z_i) \} (\mathbb{E}_{q(\bm \pi_s)}[\ln \pi_{s, s_\mathrm{ch}}] + \ln \zeta_{i,s_\mathrm{ch}}) \nonumber \\
    &\quad - \ln { \sum_{s' \in \mathrm{Ch}(s_\lambda)} \xi_{i,s_\lambda,s'}}.
\end{align}
Since the last term is independent of $\bm z_i$, this formula is equivalent to \eqref{root_is_const}. Consequently, Proposition \ref{prop:q_z} is proved.

\section{Proof of Theorem \ref{theo:q_T}}\label{appendix_q_T}

Calculating \eqref{q_star_T}, we obtain the following equation.
\begin{align}
    \ln q(T_i) &= \sum_{s \in \mathcal{L}(T_i)} \ln \phi_{i,s} + \sum_{s \in \mathcal{I}(T_i)} \ln \tilde{g}_s \nonumber \\
    &\textstyle \quad + \sum_{s \in \mathcal{L}(T_i)} \ln \tilde{g}_s^c + \mathrm{const.}, \label{q_pi_T_1}
\end{align}
where $\ln \phi_{i,s}$, $\ln \tilde{g}_s$, and $\ln \tilde{g}_s^c$ are defined in \eqref{phi_def}, \eqref{g_tilde_def}, and \eqref{g_tilde_c_def}, respectively.

Therefore, the following holds.
\begin{align}
     q(T_i) = \frac{1}{Z} \prod_{s \in \mathcal{I}(T_i)} \tilde{g}_s \prod_{s' \in \mathcal{L}(T_i)} \tilde{g}_{s'}^c \phi_{i,s'}, \label{q_T_before_reparametrization}
\end{align}
where $Z$ is a normalization term defined as follows.
\begin{align}
    Z = \sum_{T_i \in \mathcal{T}} \prod_{s \in \mathcal{I}(T_i)} \tilde{g}_s \prod_{s' \in \mathcal{L}(T_i)} \tilde{g}_{s'}^c \phi_{i,s'}.
\end{align}
Therefore, we can regard $\prod_{s \in \mathcal{L}(T_i)} \phi_{i,s}$ as a kind of likelihood and $\prod_{s \in \mathcal{I}(T_i)} \tilde{g}_s \prod_{s' \in \mathcal{L}(T_i)} \tilde{g}_{s'}^c$ as a prior distribution without a normalization term.

Next, we prove that \eqref{q_T_before_reparametrization} is equivalent to \eqref{q_T}. In other words, we reparametrize \eqref{q_T_before_reparametrization} with $\hat{g}_{i,s}$, which is defined in \eqref{g_hat} and \eqref{ln_rho}. The proof is similar to that of Theorem 7 in \cite{full_rooted_trees}. First, $Z = \rho_{i,s_\lambda}$ is straightforwardly proved by Theorem 1 in \cite{full_rooted_trees}, where $\rho_{i,s}$ is defined in \eqref{ln_rho}.

Next, we substitute the definitions of $\hat{g}_{i,s}$ and $\rho_{i,s}$ into \eqref{q_T} and bring it back to \eqref{q_T_before_reparametrization}.
\begin{align}
    &\prod_{s \in \mathcal{I}(T_i)} \hat{g}_{i,s} \prod_{s' \in \mathcal{L}(T_i)} (1-\hat{g}_{i,s'}) \nonumber \\
    &= \prod_{s \in \mathcal{I}(T_i)} \hat{g}_{i,s} \prod_{s \in \mathcal{L}(T_i) \setminus \mathcal{L}_\mathrm{max}}  (1-\hat{g}_{i,s})  \prod_{s \in \mathcal{L}(T_i) \cap \mathcal{L}_\mathrm{max}}  1 \\
    &\overset{(a)}{=} \prod_{s \in \mathcal{I}(T_i)} \frac{\tilde{g}_s \prod_{s' \in \mathrm{Ch}(s)} \rho_{i,s'}}{\rho_{i,s}} \nonumber \\
    &\qquad \times \prod_{s \in \mathcal{L}(T_i) \setminus \mathcal{L}_\mathrm{max}}   \left(  1  -  \frac{\tilde{g}_s  \prod_{s' \in \mathrm{Ch}(s)}  \rho_{i,s'}}{\rho_{i,s}}  \right) \nonumber \\
    &\qquad \times \prod_{s \in \mathcal{L}(T_i) \cap \mathcal{L}_\mathrm{max}} \frac{\tilde{g}_s^c \rho_{i,s}}{\rho_{i,s}}  \\  
    &= \prod_{s \in \mathcal{I}(T_i)} \frac{\tilde{g}_s \prod_{s' \in \mathrm{Ch}(s)} \rho_{i,s'}}{\rho_{i,s}} \nonumber \\
    &\qquad \times \prod_{s \in \mathcal{L}(T_i) \setminus \mathcal{L}_\mathrm{max}}   \frac{\rho_{i,s}   -  \tilde{g}_s  \prod_{s' \in \mathrm{Ch}(s)}  \rho_{i,s'}}{\rho_{i,s}} \nonumber \\
    &\qquad \times   \prod_{s \in \mathcal{L}(T_i) \cap \mathcal{L}_\mathrm{max}}  \frac{\tilde{g}_s^c \phi_{i,s}}{\rho_{i,s}} \\  
    &= \prod_{s \in \mathcal{I}(T_i)} \frac{\tilde{g}_s \prod_{s' \in \mathrm{Ch}(s)} \rho_{i,s'}}{\rho_{i,s}} \nonumber \\
    &\qquad \times \prod_{s \in \mathcal{L}(T_i) \setminus \mathcal{L}_\mathrm{max}} \frac{\tilde{g}_s^c \phi_{i,s}}{\rho_{i,s}} \prod_{s \in \mathcal{L}(T_i) \cap \mathcal{L}_\mathrm{max}} \frac{\tilde{g}_s^c \phi_{i,s}}{\rho_{i,s}} \\  
    &= \prod_{s \in \mathcal{I}(T_i)} \frac{\tilde{g}_s \prod_{s' \in \mathrm{Ch}(s)} \rho_{i,s'}}{\rho_{i,s}} \prod_{s \in \mathcal{L}(T_i)} \frac{\tilde{g}_s^c \phi_{i,s}}{\rho_{i,s}} \label{telescope} \\
    &= \frac{1}{\rho_{i,s_\lambda}}  \prod_{s \in \mathcal{I}(T_i)} \tilde{g}_s \prod_{s' \in \mathcal{L}(T_i)} \tilde{g}_{s'}^c \phi_{i,s'} = \eqref{q_T_before_reparametrization},
\end{align}
where $(a)$ is due to $\tilde{g}_s^c = 1$ for $s \in \mathcal{L}_\mathrm{max}$.
Here, \eqref{telescope} is a telescoping product, i.e., $\rho_{i,s}$ appears at once in each of the denominator and the numerator. Therefore, we can cancel them except for $\rho_{i,s_\lambda}$. Consequently, \eqref{q_T_before_reparametrization} is equivalent to \eqref{q_T}. It should be noted that we did not make any approximations to derive \eqref{q_T} from \eqref{q_star_T}. \hfill $\Box$

\end{document}